\documentclass[10pt,letterpaper]{article}
\title{Deep Learning Predicts Hip Fracture using Confounding Patient and Healthcare Variables}

\usepackage[top=0.85in,left=0.85in,footskip=0.85in]{geometry}
\usepackage{amsmath,amssymb}  % amsmath and amssymb packages, useful for mathematical formulas and symbols
\usepackage{changepage}  % Use adjustwidth environment to exceed column width (see example table in text)
\usepackage[utf8x]{inputenc}  % Use Unicode characters when possible
\usepackage{textcomp,marvosym}  % textcomp package and marvosym package for additional characters
\usepackage[superscript]{cite}  % cite package, to clean up citations in the main text. Do not remove.
\usepackage{nameref}  % Use nameref to cite supporting information files (see Supporting Information section for more info)
\usepackage{microtype}  % ligatures disabled
\DisableLigatures[f]{encoding = *, family = * }
\usepackage[table]{xcolor}  % color can be used to apply background shading to table cells only
\usepackage{array}  % array package and thick rules for tables
\newcolumntype{+}{!{\vrule width 2pt}}  % create "+" rule type for thick vertical lines

\usepackage{multirow}
\usepackage[normalem]{ulem}
\useunder{\uline}{\ul}{}

\usepackage{graphicx}
\usepackage{subcaption}

\usepackage{float}  %for float
\usepackage{wrapfig}  %for wrap figures

\usepackage[colorinlistoftodos]{todonotes}  % for comments

\newlength\savedwidth

\raggedright
\usepackage[aboveskip=1pt,font=small,labelfont=bf,labelsep=period,justification=raggedright,singlelinecheck=off]{caption}

\makeatletter
\renewcommand{\@biblabel}[1]{\quad#1.}
\makeatother
\usepackage{lastpage,fancyhdr}  % Header and Footer with logo
\usepackage{epstopdf}  % generate low resolution images

\fancyheadoffset[L]{2.25in}
\fancyfootoffset[L]{2.25in}
\usepackage{longtable}
\begin{document}
\vspace*{0.2in}

\begin{flushleft}
{\Large
\textbf\newline{Deep Learning Predicts Hip Fracture using Confounding Patient and Healthcare Variables}
}
\newline

Marcus A. Badgeley\textsuperscript{1,2,3},
John R. Zech\textsuperscript{4},
Luke Oakden-Rayner\textsuperscript{5},
Benjamin S. Glicksberg\textsuperscript{6},
Manway Liu\textsuperscript{1},
William Gale\textsuperscript{7},
Michael V. McConnell\textsuperscript{1,8},
Beth Percha\textsuperscript{2},
Thomas M. Snyder\textsuperscript{1},
Joel T. Dudley\textsuperscript{2,3}
\\
\bigskip

\textbf{1} Verily Life Sciences LLC, South San Francisco, CA, USA
\\\textbf{2} Institute for Next Generation Healthcare, Icahn School of Medicine at Mount Sinai, New York, NY, USA
\\\textbf{3} Department of Genetics and Genomic Sciences, Icahn School of Medicine at Mount Sinai, New York, NY, USA
\\\textbf{4} Department of Medicine, California Pacific Medical Center, San Francisco, CA, USA
\\\textbf{5} School of Public Health, The University of Adelaide, Adelaide, South Australia
\\\textbf{6} Bakar Computational Health Sciences Institute, University of California, San Francisco, CA, USA
\\\textbf{7} School of Computer Sciences, The University of Adelaide, Adelaide, South Australia
\\\textbf{8} Division of Cardiovascular Medicine, Stanford School of Medicine, Stanford, CA, USA
\bigskip

\end{flushleft}

\section*{Abstract}
Hip fractures are a leading cause of death and disability among older adults. Hip fractures are
also the most commonly missed diagnosis on pelvic radiographs, and delayed diagnosis leads
to higher cost and worse outcomes. Computer-Aided Diagnosis (CAD) algorithms have shown
promise for helping radiologists detect fractures, but the image features underpinning their
predictions are notoriously difficult to understand. In this study, we trained deep learning
models on 17,587 radiographs to classify fracture, five patient traits, and 14 hospital process
variables. All 20 variables could be predicted from a radiograph (p \textless 0.05), with the best
performances on scanner model (AUC=1.00), scanner brand (AUC=0.98), and whether the
order was marked “priority” (AUC=0.79). Fracture was predicted moderately well from the
image (AUC=0.78) and better when combining image features with patient data (AUC=0.86,
p=2e-9) or patient data plus hospital process features (AUC=0.91, p=1e-21). The model
performance on a test set with matched patient variables was significantly lower than a random
test set (AUC=0.67, p=0.003); and when the test set was matched on patient and image
acquisition variables, the model performed randomly (AUC=0.52, 95\% CI 0.46-0.58), indicating
that these variables were the main source of the model’s predictive ability overall. We also used
Naive Bayes to combine evidence from image models with patient and hospital data and found
their inclusion improved performance, but that this approach was nevertheless inferior to directly
modeling all variables. If CAD algorithms are inexplicably leveraging patient and process
variables in their predictions, it is unclear how radiologists should interpret their predictions in
the context of other known patient data. Further research is needed to illuminate deep learning
decision processes so that computers and clinicians can effectively cooperate.

\section*{Introduction}
\noindent An estimated 1.3 million hip fractures occur annually and are associated with 740,000 deaths
and 1.75 million disability adjusted life-years.\cite{Johnell2004-py} The chance of death in the three months following
a hip fracture increases by fivefold for women and eightfold for men, relative to age- and sex-
matched controls.\cite{Haentjens2010-ig} When a middle-aged or elderly patient presents with acute hip pain and
fracture is suspected, clinical guidelines recommend first ordering a hip radiograph.\cite{Ward2013-kr} However,
not all fractures are detectable on radiographs.\cite{Cannon2009-ht, Kirby2010-mu} If a patient with high clinical suspicion of
fracture has a negative or indeterminant radiograph, then it is usually appropriate to follow-up
with a pelvic MRI.\cite{Ward2013-kr} Fractures are the most commonly missed diagnosis on radiographs of the
spine and extremities, and the majority of these errors are perceptual (i.e., a radiologist not
noticing some abnormality as opposed to misinterpreting a recognized anomaly).\cite{Donald2012-rr}
\bigskip

\noindent Statistical learning models can both detect fractures and help radiologists detect fractures.  Past studies used machine learning (ML) to identify combinations of hand-engineered features associated with fracture, and more recent studies used deep learning (DL) to discover hierarchical pixel patterns from many images with a known diagnosis.  Most studies detect fracture in algorithm-only systems.\cite{Chai2011-ft, Donnelley2008-hh, Kazi2017-sd} Kazai et al. performed a clinical trial to study how algorithms can augment radiologists and found radiologists were significantly better at detecting vertebral fractures when aided by an ML model that had a standalone sensitivity of 81\%.\cite{Kasai2008-vq} Convolutional Neural Networks (CNNs), the DL models best suited for image recognition, have recently been used to detect fracture in the appendicular skeleton including wrists\cite{Kim2018-zf}, shoulders\cite{Chung2018-uw}, and hands and feet\cite{Olczak2017-ob}.  Gale et al. developed the only previously-reported hip fracture detector using DL; their model achieved an area under the receiver operating curve (AUC) of 0.994.\cite{Gale2017-ae} These academic DL reports compared isolated image model performance against humans, but none tested whether algorithms could aid human diagnosis.  In contrast, the company Imagen Technologies’ OsteoDetect DL system reported improving humans from an unaided AUC 0.84 to AUC 0.89, according to a letter from the FDA (https://www.accessdata.fda.gov/cdrh\_docs/pdf18/DEN180005.pdf).  Deep learning studies on various image-based fracture prediction tasks have been published, but they do not consider patient and hospital covariates or how algorithms can augment human decision processes.  
\bigskip

\noindent Statistical learning algorithms sometimes learn unintended or unhelpful patterns contained in
the model training data. Diverse examples for common DL applications include gender being
differentially classified on photographs of the face depending on a person’s race\cite{Buolamwini2018-pd} as well as
gender detection on photographs of the outer eye leveraging disproportionate use of mascara.\cite{Kuehlkamp2017-yc} 
Language processing algorithms learned to perpetuate human prejudices from text from the
Web.\cite{Caliskan2016-el} In medicine, retrospectively collected observational datasets may have patient and
healthcare process biases. An ML study on medical record data found that hospital process
variables were more predictive of patient mortality than biological signal.\cite{Agniel2018-qu} Statistical learning
can exploit biases that are pervasive in observational human datasets.
\bigskip

\noindent DL has previously been shown to detect patient demographics and clinical variables from fundoscopy images.\cite{Poplin2018-wz} We previously showed that DL can learn individual hospital sources in a multi-site trial and leverage this for disease detection, which leads to inconsistent performance when deployed to new hospitals.\cite{Zech2018-ok} Here we perform the first comprehensive analysis of what patient and hospital process variables DL can detect in radiographs and whether they contribute to the inner workings of a fracture detection model. We group variables into disease (fracture), patient (sex, age, body mass index, pain, and recent fall), and hospital processes (e.g, department, device model, study date).  We use patient and hospital process data to develop multimodal models and to create matched patient cohorts to test image-only models. To assess the suitability for image-only models augmenting human interpretations, we experiment with Naive Bayes model ensembles.  We re-analyze test data from the best published hip fracture model by Gale et al. and conclude by highlighting design of clinical experiments and strategies to mitigate the susceptibility of models to confounding variables.

\begin{figure}
\centering
\includegraphics[width=0.66\textwidth]{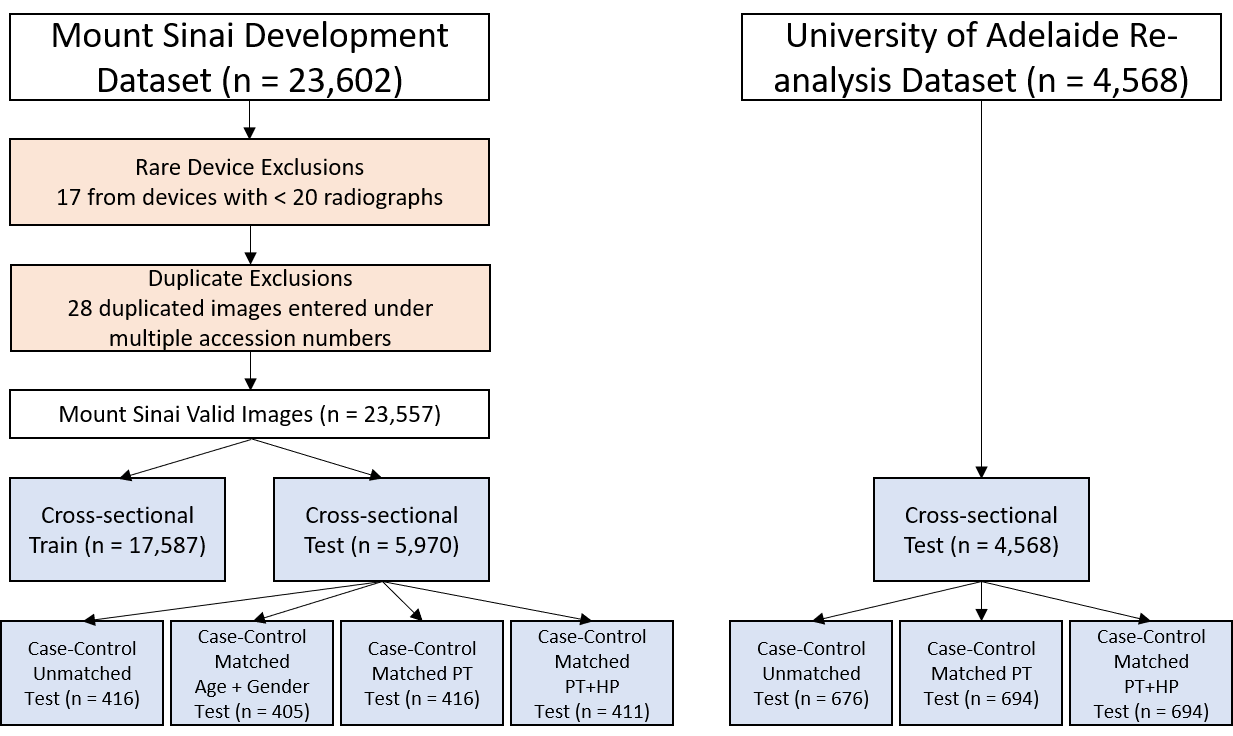}
\caption{Cohort Waterfall Schematic with Preprocessing Exclusions and Subsampling.}
\label{fig:FigS1}
\end{figure}

\section*{Results}

\subsection*{Dataset and Unsupervised Analysis}
\begin{figure}
\centering
\includegraphics[width=1\textwidth]{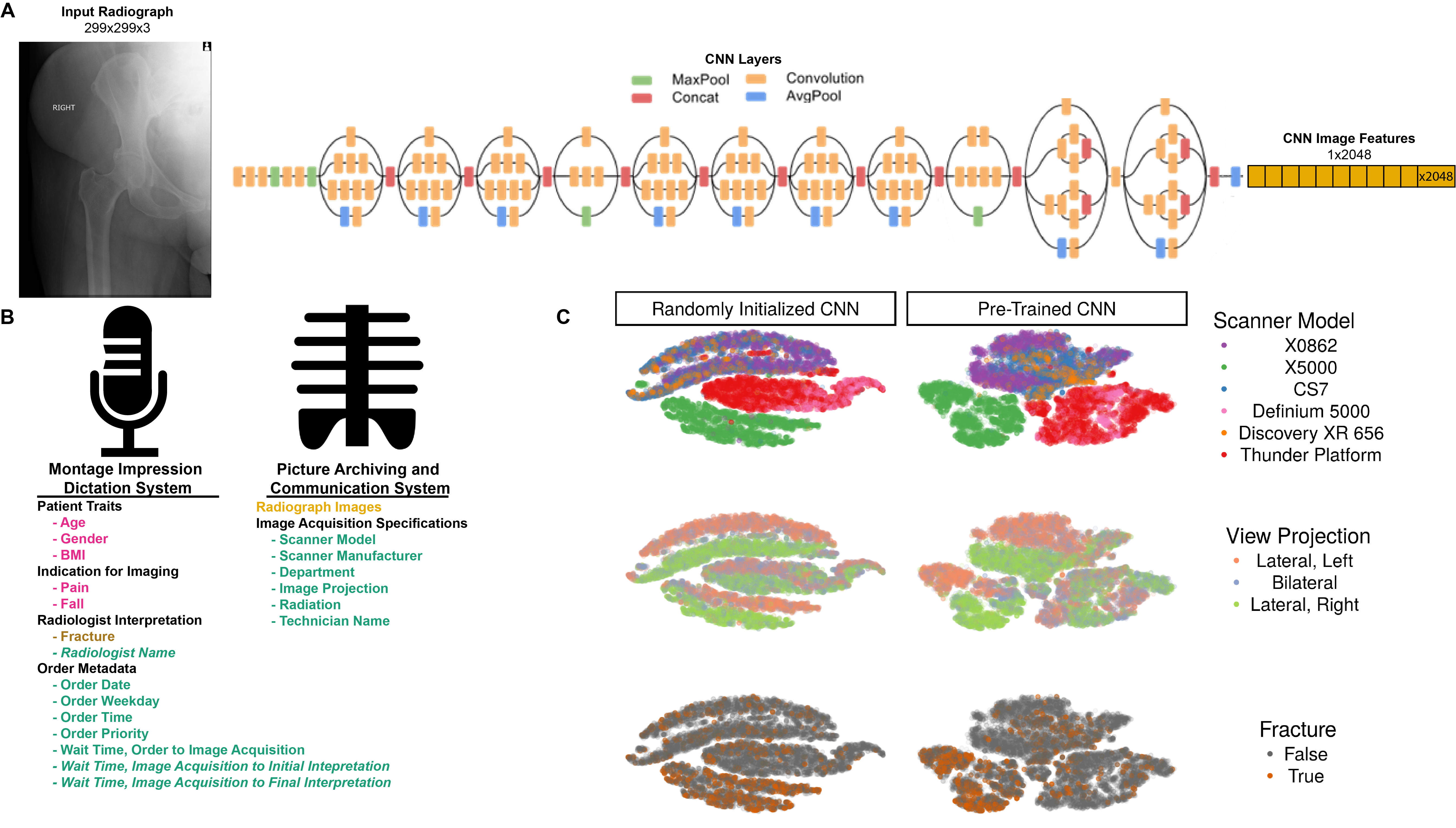}
\caption{The Main Source of Variation in Whole Radiographs is Explained by the Device used to Capture the Radiograph. A) Schematic of the inception v-3 deep learning model used to featurize radiographs into an embedded 2048-dimensional representation.  Inception model architecture schematic derived from https://cloud.google.com/tpu/docs/inception-v3-advanced.  B) Data were collected from two sources.  Variables were categorized as pathology (gold), image (IMG, yellow), patient, (PT, pink), or hospital process (HP, green).  Only variables known at the time of image acquisition were used as explanatory variables.  C) The distribution of radiographs projected into clusters by t-Distributed Stochastic Neighbor Embedding (t-SNE) and coloring is applied independently of position to designate how the unsupervised distribution of clusters happens to relate to hip fracture and categorical variables.}
\label{fig:Fig1}
\end{figure}

\begin{figure}
\centering
\includegraphics[width=0.5\textwidth]{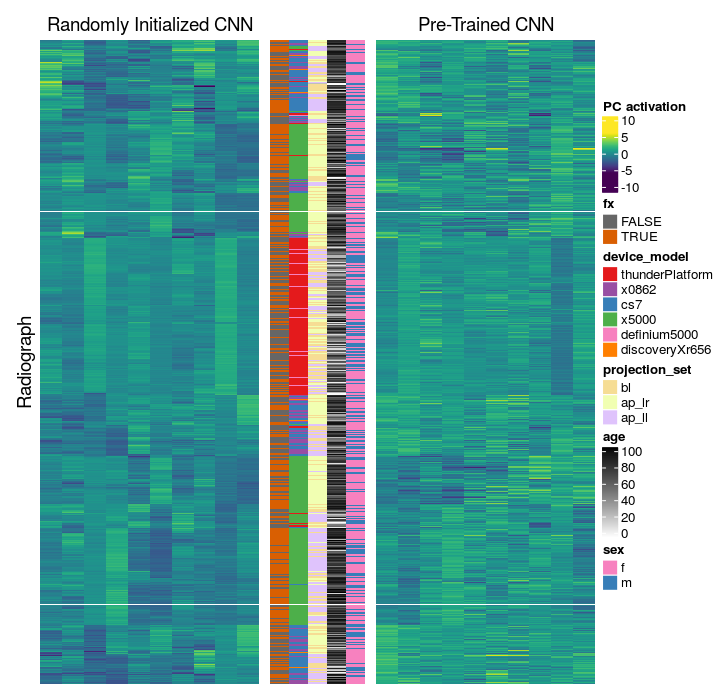}
\caption{Image Feature Matrix annotated with fracture and covariates.  The image training data is represented by a row for each sample radiograph and a column for each CNN principle component feature.  The fill reflects the neural activation of each feature for each radiograph.  Radiographs are clustered and annotated with fracture and several covariates.  For this figure samples were enriched for fracture by randomly sampling 500 images with and 500 without fracture.}
\label{fig:FigS2}
\end{figure}

We collected 23,602 hip radiographs and associated patient and hospital process data from the medical imaging and clinician dictation databases, of which 23,557 were used to train and test (3:1 split) Convolutional Neural Networks (CNNs) (Figure ~\ref{fig:FigS1}).  Extracted features are separated into disease (i.e., fracture), image (IMG), patient (PT), or hospital process (HP) variables, and only those known at the time of image acquisition are used as explanatory variables (Figure ~\ref{fig:Fig1}B). We used the inception-v3 model architecture (Figure ~\ref{fig:Fig1}A) and computed image features for all radiographs with randomly initialized model weights and weights pre-trained on everyday images.  The pre-trained model takes a 299×299 pixel input and computes 2048 8×8 feature maps, and we averaged each feature map to get a 2048-dimensional feature vector.  Clustering analyses showed that the greatest source of variation between radiographs is the scanner that captured the image, and within each scanner the projection view forms further discrete clusters (Figure ~\ref{fig:Fig1}C).  The image feature matrix demonstrates a concomitant clustering by these image acquisition features (Figure ~\ref{fig:FigS2}).  The scanner model is the best predictor of the first Principal Component (PC) (R$^2$ = 0.59) and 8 of the first 10 PCs.

\subsection*{Modeling Fracture, Patient Traits, and Hospital Process Variables}
\begin{figure}
\centering
\includegraphics[width=1\textwidth]{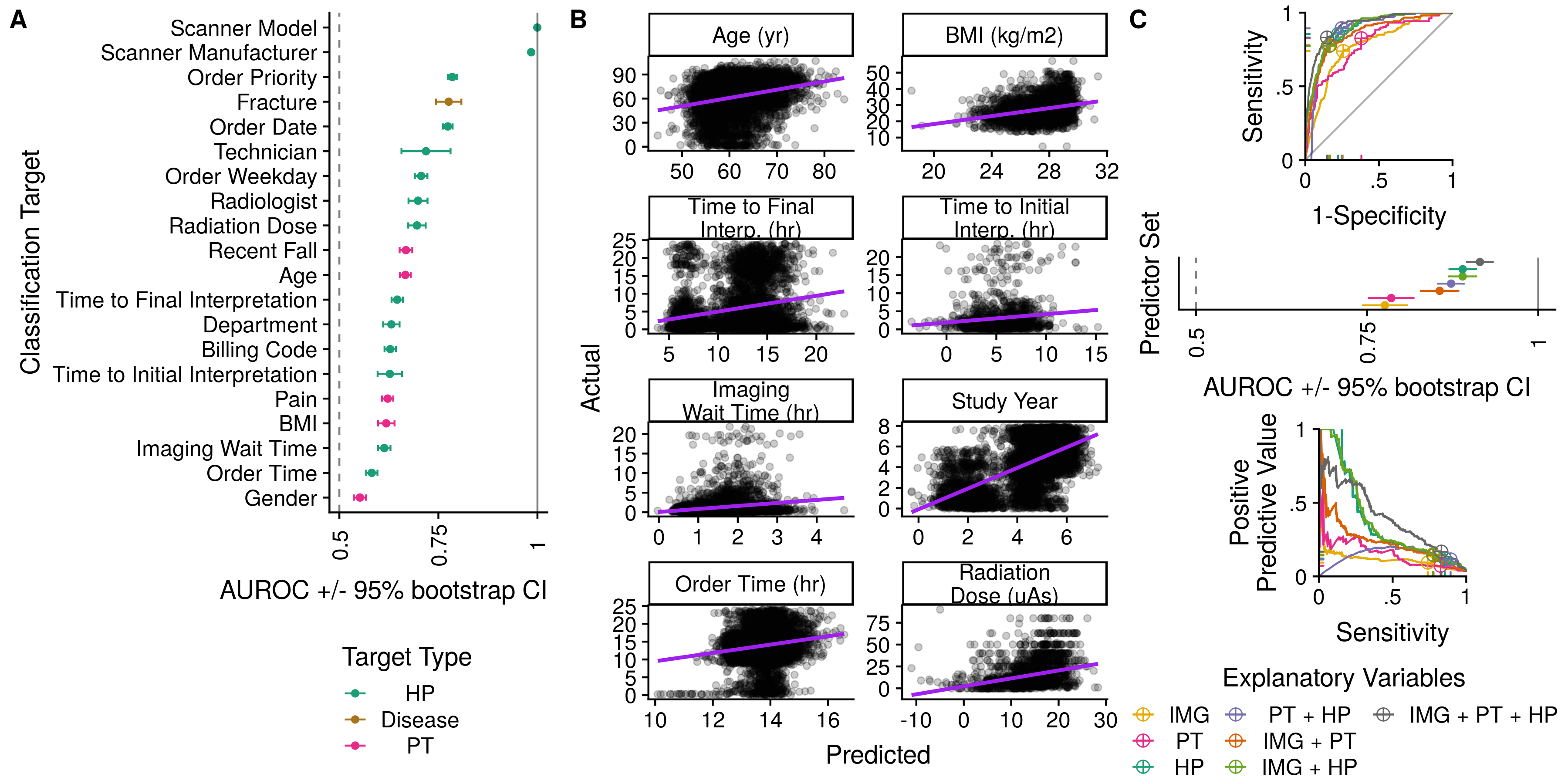}
\caption{Deep Learning Predicts All Patient and Hospital Processes from a Radiograph, and All Patient and Hospital Process Factors are Associated with Fracture.  A) Deep learning image models to predict binarized forms of 14 HP variables, 5 PT variables, and hip fracture.  Error bars indicate the 95\% confidence intervals of 2000 bootstrapped samples.  B) Deep learning regression models to predict 8 continuous variables from hip radiographs.  Each dot represents one radiograph, and the purple lines are linear models of actual versus predicted values.  C) ROC, ROC +/- bootstrap confidence intervals, and Precision Recall Curves for deep learning models that predict fracture based on combinatorial predictor sets of IMG, PT, and HP variables.  Crosshairs indicate the best operating point on ROC and PRC curves.}
\label{fig:Fig2}
\end{figure}

\begin{wrapfigure}[22]{R}{0.5\textwidth}
\centering
\includegraphics[width=0.5\textwidth]{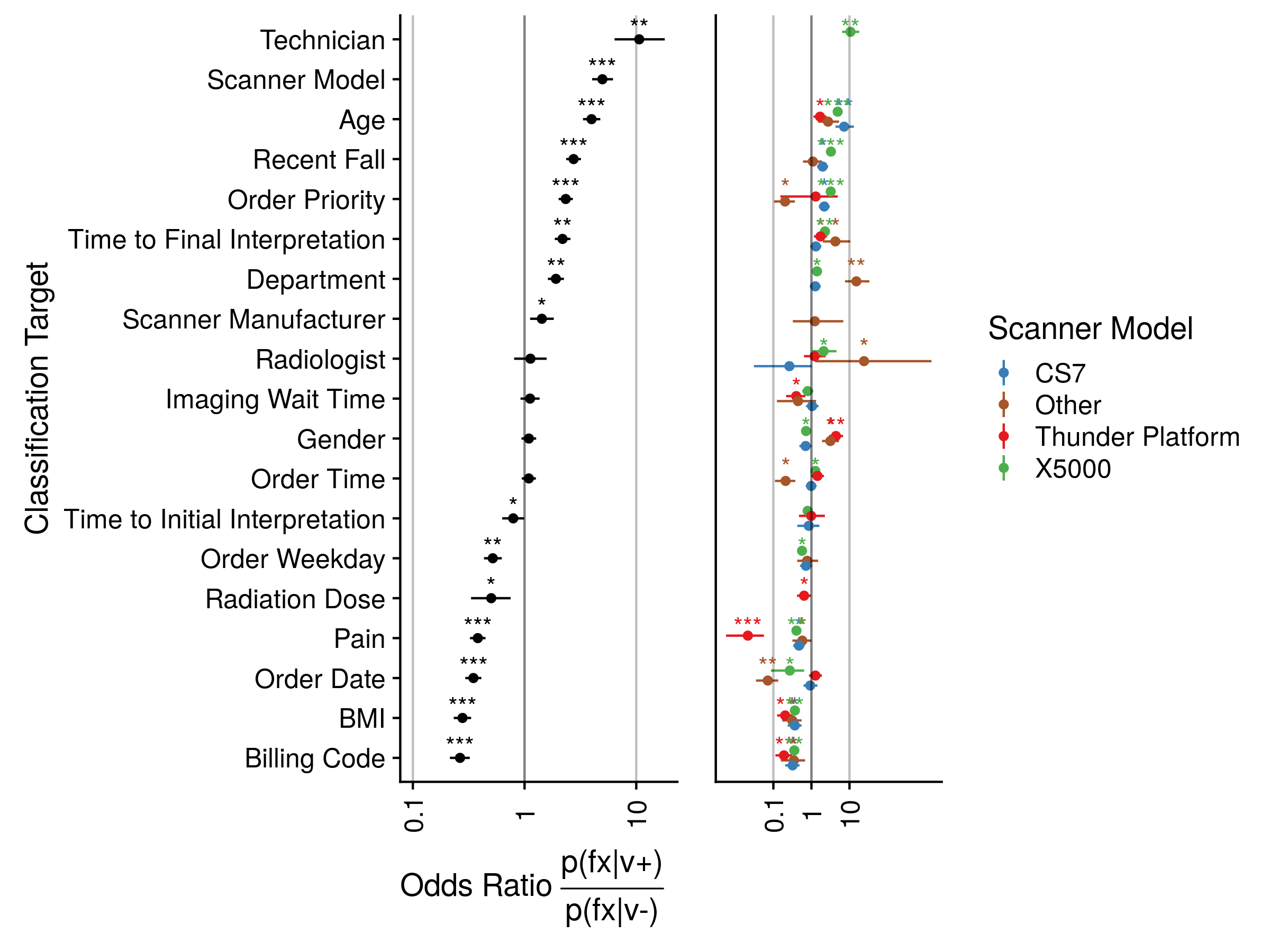}
\caption{Association between fracture and covariates.  Univariate associations between hip fracture and each covariate were assessed using Fisher’s Exact test on the full dataset (left) and after stratifying by the scanner device (right).  Each covariate was binarized as described in the supplemental methods.  Significance indicators: * = p\textless0.05, ** = p\textless1e-10, and *** = p\textless1e-25.}
\label{fig:FigS5}
\end{wrapfigure}

We transformed all scalar variables into binary factors and trained logistic regression models for fracture, PTs, and HPs as described in detail in the methods section.  All 20 of 20 image models were significantly better than random (p \textless 0.05) (Figure \ref{fig:Fig2}A).  Hip fracture was detected with AUC 0.78 (95\% CI: 0.74-0.81), and the best detected secondary targets were the device that took the scan (AUC 1, CI 1-1), scanner manufacturer (AUC 0.98, 95\% CI 0.98-0.99), and whether the image was ordered as high priority (AUC 0.79, 95\% CI 0.77-0.80).  The difference in performance across targets is not explained by differences in total sample size or the number of examples in the smaller class.  Most of the patient and hospital process factors (15/19) were themselves significantly associated with fracture (Fisher’s exact test, p \textless 0.05); and, after stratifying by device, the other four covariates were associated with fracture on at least one device (Figure \ref{fig:FigS5}).  The best predicted continuous variable is the year the scan was ordered (R$^2$ = .39) (Figure \ref{fig:Fig2}B).

\subsection*{Multimodal Deep Learning Models}
We then compared combinatorial sets of IMG, PT, and HP features for fracture prediction.  Missing PT+HP data was imputed as described in the supplementary methods.  Multivariate model performance metrics are provided in Supplementary Table 3 and operating point independent statistics are shown in Figure ~\ref{fig:Fig2}C.  Fracture was better predicted by HP features (AUC 0.89, 95\% CI 0.87-0.91) than either IMG features (AUC 0.78, p=3e-17) or PT features (AUC 0.79, p=9e-12).  Adding IMG to the HP set did not improve performance (p=0.97).  The best predictor set was the full set of IMG + PT + HP (AUC 0.91, 95\% CI 0.90-0.93).

\subsection*{Evaluating CNNs on Matched Patient Populations}
\begin{figure}
\centering
\includegraphics[width=1\textwidth]{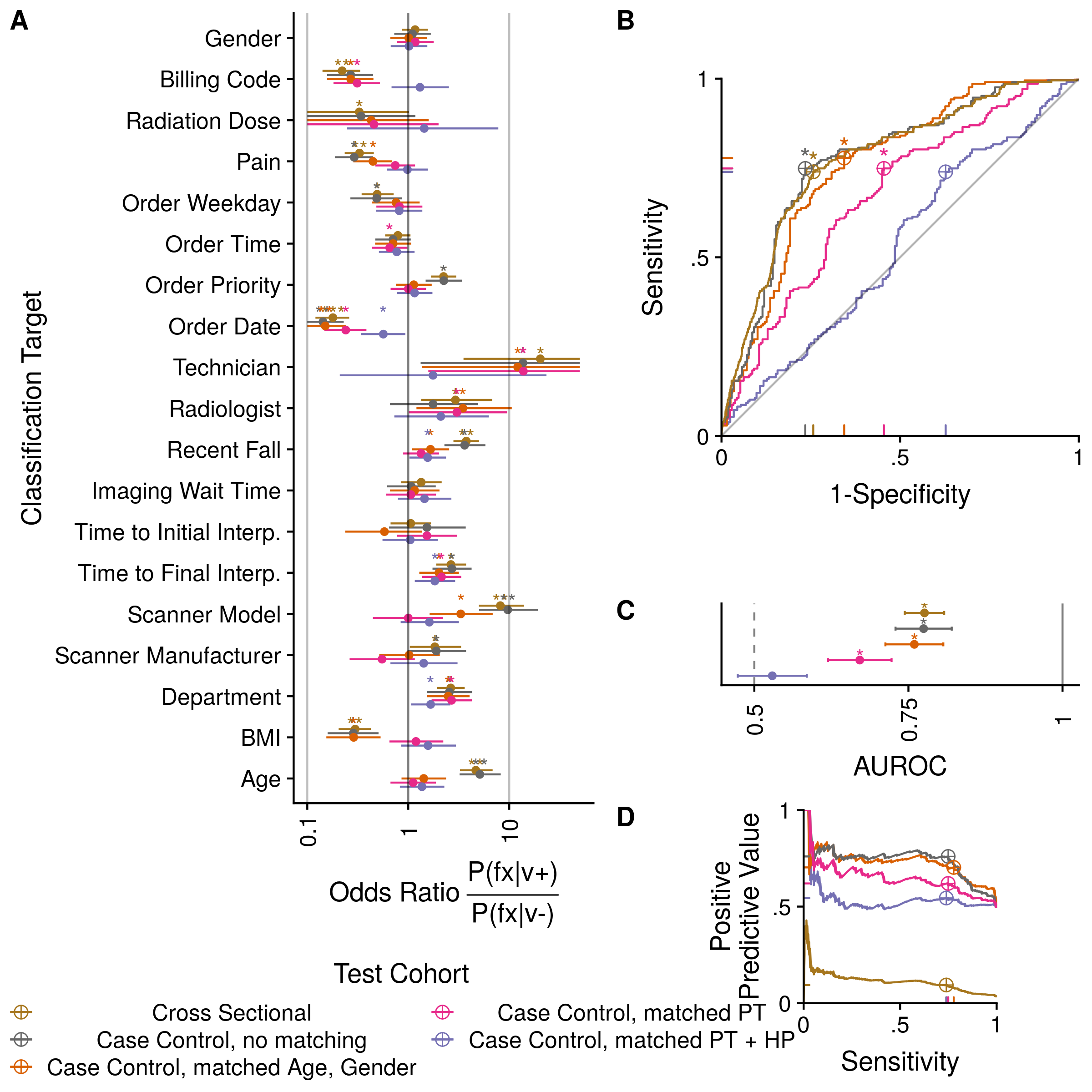}
\caption{Deep learning hip fracture from radiographs is successful until controlling for all patient and hospital process variables.  A) The association between each metadata variable and fracture, colored by how the test cohort is sampled.  (*) indicate a Fisher’s Exact test with p\textless0.05.  B) ROC and D) Precision Recall curves for the image-classifier tested on differentially sampled test sets.  The best operating point is indicated with crosshairs.  (*) represents a 95\% confidence interval that does not include 0.5.  C) Summary of (B) with 95\% bootstrap confidence intervals.}
\label{fig:Fig3}
\end{figure}

We sought to disentangle the ability of a CNN to directly detect fracture versus indirectly predicting fracture by detecting confounding variables associated with fracture. We manipulate the associations between confounders and fracture by subsampling the full (cross-sectional) dataset in a case-control fashion with variable matching routines.  To control for the decreased sample size while maintaining the natural fracture-covariate associations, we downsampled the non-fracture images by randomly selecting one normal radiograph for each fracture in the test set.  For matched subsets, we computed a distance metric between each case and all controls as described in the methods and derived cohorts matched on demographics (age, gender), PT variables or PT+HP variables (Table \ref{Table1}).  With increasingly comprehensive matching, the number of fracture-associated covariates consistently decreased (Figure \ref{fig:Fig3}A).
\bigskip

\begin{table}
\centering
\caption{ Cohort Characteristics after various Sampling Routines.  The full cross-sectional dataset was divided into train and test sets, and the test set was further subsampled in a case-control fashion to generate several balanced test sets.}
\label{Table1}
\resizebox{0.92\textwidth}{!}{\begin{tabular}{|l|l|l|l|l|l|l|}
\hline
Cohort & cs-train & cs-test & cc-rnd-test & cc-dem-test & cc-pt-test & cc-pthp-test\\
\hline
Sampling & Cross-Sectional & Cross-Sectional & Case-Control & Case-Control & Case-Control & Case-Control\\
\hline
Matching & NA & NA & Random & AgeGender & Pt & PtHp\\
\hline
Partition & train & test & test & test & test & test\\
\hline
No. radiographs & 17,587 & 5,970 & 416 & 405 & 416 & 411\\
\hline
No. patients & 6,768 & 2,256 & 275 & 252 & 217 & 186\\
\hline
No. scanners & 11 & 11 & 10 & 9 & 8 & 6\\
\hline
No. scanner manufacturers & 4 & 4 & 4 & 4 & 4 & 4\\
\hline
Age, mean (SD), years & 61 (22) & 61 (22) & 67 (24) & 75 (20) & 75 (21) & 74 (19)\\
\hline
Female frequency, No. (\%) & 11,647 (66) & 3,873 (65) & 260 (62) & 249 (61) & 263 (63) & 253 (62)\\
\hline
Fracture frequency, No. (\%) & 572 (3) & 207 (3) & 207 (50) & 207 (51) & 207 (50) & 207 (50)\\
\hline
BMI, mean (SD) & 28 (7) & 28 (7) & 25 (5) & 25 (5) & 24 (5) & 24 (4)\\
\hline
Fall frequency, No. (\%) & 3,214 (18) & 1,139 (19) & 133 (32) & 160 (40) & 174 (42) & 165 (40)\\
\hline
Pain frequency, No. (\%) & 9,010 (51) & 2,960 (50) & 164 (39) & 137 (34) & 117 (28) & 104 (25)\\
\hline
\end{tabular}}
\end{table}

\noindent We then evaluated the image-only classifier for fracture on each test set (Figure \ref{fig:Fig3}B-D, Table \ref{TableS6}).  The area under the Precision Recall Curve (PRC) is dependent on the disease prevalence, and since the original population had a 3\% fracture prevalence but case-control cohorts have a 50\% prevalence, the PRC is significantly higher for case-control cohorts.  Random subsampling had no effect on the primary evaluation metric, AUC (0.78 vs 0.77, p=0.96).  Model performance was consistent after matching by demographics (AUC=0.76, p=0.65) but significantly lower after matching by all patient variables (AUC=0.67, p=0.003).  When evaluated on a test cohort matched by all covariates, the fracture detector was no longer better than random (AUC=0.52, 95\% CI 0.46-0.58) and significantly worse than when assessed on all other test cohorts.

\begin{table}
\centering
\caption{Performance of an image-based fracture detection model evaluated on test-sets with variable case-control sampling strategies.}
\label{TableS6}
\resizebox{0.92\textwidth}{!}{\begin{tabular}{|l|l|l|l|l|l|l|l|l|l|l|l|l|}
\hline
Test Cohort & auc & auprc & threshold & specificity & sensitivity & accuracy & npv & ppv & tn & tp & fn & fp\\
\hline
Cross Sectional & 0.78 (0.74-0.81) & 0.11 & 0.033 & 0.74 & 0.74 & 0.74 & 0.99 & 0.094 & 4,283 & 153 & 54 & 1,480\\
\hline
Case Control, no matching & 0.77 (0.73-0.82) & 0.74 & 0.032 & 0.77 & 0.75 & 0.76 & 0.75 & 0.760 & 160 & 155 & 52 & 49\\
\hline
Case Control, matched Age, Gender & 0.76 (0.71-0.81) & 0.73 & 0.029 & 0.66 & 0.78 & 0.72 & 0.74 & 0.703 & 130 & 161 & 46 & 68\\
\hline
Case Control, matched PT & 0.67 (0.62-0.72) & 0.65 & 0.032 & 0.55 & 0.75 & 0.65 & 0.69 & 0.620 & 114 & 155 & 52 & 95\\
\hline
Case Control, matched PT + HP & 0.53 (0.47-0.59) & 0.54 & 0.033 & 0.37 & 0.74 & 0.56 & 0.58 & 0.544 & 76 & 153 & 54 & 128\\
\hline
\end{tabular}}
\end{table}

\subsection*{Evaluating Effect of Confounding Variables on Model of Gale et al.}
\begin{figure}
\centering
\includegraphics[width=0.66\textwidth]{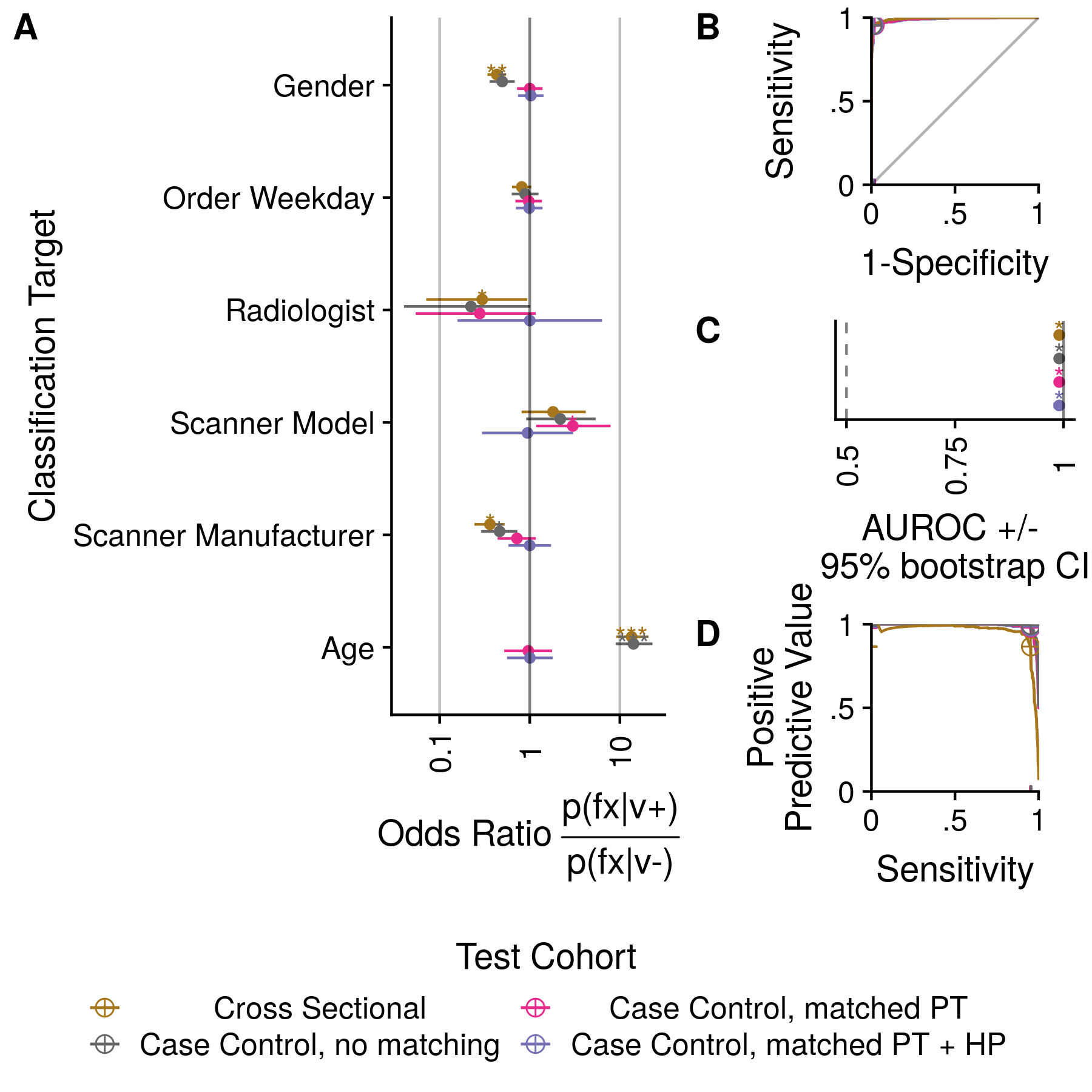}
\caption{Association of covariates and fracture and the performance of fracture detection models evaluated on differentially sampled test cohorts from the Adelaide dataset.  A) The association between each covariate and fracture, colored by how the test cohort is sampled.  (*) indicate a Fisher’s Exact test with p<0.05.  B) ROC and D) Precision Recall curves for the image-classifier tested on differentially sampled test sets.  The best operating point is indicated with crosshairs.  (*) represents a 95\% confidence interval that does not include 0.5.  C) Summary of (B) with 95\% bootstrap confidence intervals.}
\label{fig:FigS6}
\end{figure}

We repeated this case-control testing experiment on the model previously reported by Gale et al\cite{Gale2017-ae}. There were fewer covariates to match patients by (Figure \ref{fig:FigS6}A), and the fracture detection results were robust across case-control subsampling routines (Figure \ref{fig:FigS6}B-D).  Performance on a randomly subsampled test (AUC 0.99, 95\% CI 0.99-1.0) was similar (p=0.14) to the fully matched cohort (AUC 0.99, 95\% CI 0.98-1.0).

\subsection*{Secondary Evidence Integration from Image Models and Clinical Data}
\begin{figure}
\centering
\includegraphics[width=1\textwidth]{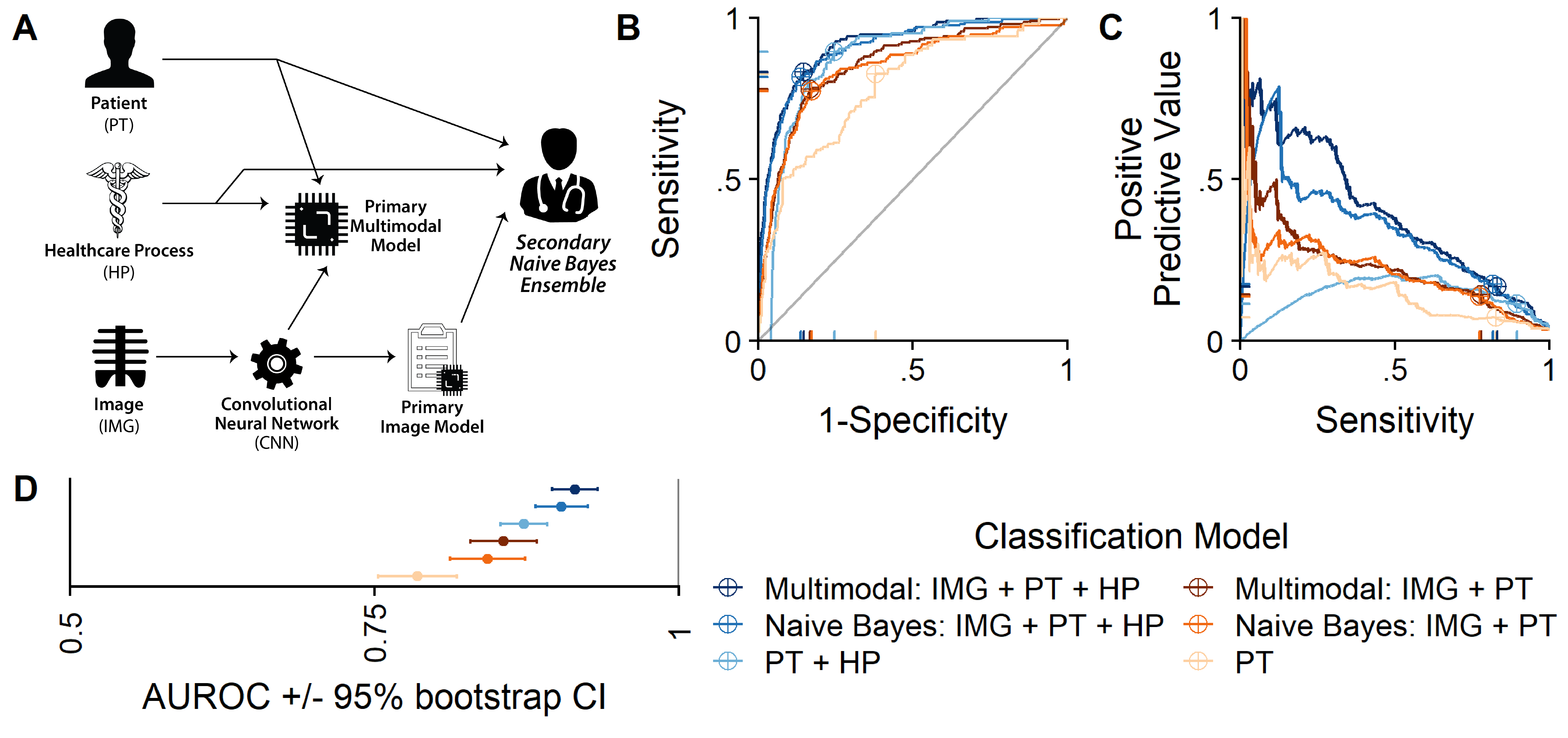}
\caption{Deep learning a compendium of patient data by direct multimodal models, or by
secondarily ensembling image-only models with other PT and HP variables. A)
experiment schematic demonstrating the CAD simulation scenario wherein a physician
secondarily integrates image-only and other clinical data (as modeled in a Naive Bayes
ensemble). B) ROC and C) Precision Recall curves for classifiers tested on differentially
sampled test sets. The best operating point is indicated with crosshairs. D) Summary of (B)
with 95\% bootstrap confidence intervals.}
\label{fig:Fig4}
\end{figure}

Computer-Aided Diagnosis (CAD) tools provide clinicians with supplemental information to make diagnostic decisions.  We have shown that deep learning models benefit from leveraging statistical relationships between fracture and patient and hospital process variables when the algorithms are tested in isolation, but we have not studied the impact this has under a CAD use scenario.  We simulate a CAD scenario by training separate models that predict fracture with individual sets of predictors and then train ensembles or multimodal models that use a combination of predictor sets (Figure \ref{fig:Fig4}A).  To simulate a clinician’s reasoning, we use a Naive Bayes ensemble that integrates the CNN’s image prediction with the likelihood of disease based on covariates. We opine that this ensemble model is reflective of how a clinician might use an image model -- without any knowledge of the inner workings of predictive models, a clinician would be unable to control for the fact that available models may be basing their predictions on same information.  For a positive control, we train a multimodal model which encodes the interdependencies between image and covariate predictor sets.
\bigskip

\noindent Multimodal models trained directly on IMG+PT and IMG+PT+HP outperform models without image data (Figure \ref{fig:Fig4}B-D).  Secondary integration of IMG predictions with PT predictions (Naive Bayes AUC 0.84, 95\% CI 0.81-0.87) is better (p = 2e-8) than considering only PT (AUC 0.79, 95\% CI 0.75-0.82), but worse (p = 0.01) than directly combining IMG+PT data (multimodal AUC 0.86, 95\% CI 0.83-0.89).  Similarly, secondary integration of IMG predictions with PT+HP predictions (Naive Bayes AUC 0.90, 95\% CI 0.88-0.93) is better (p = 5e-11) than considering only PT+HP (AUC 0.87, 95\% CI 0.85-0.89) but worse (p = 0.004) than directly combining IMG+PT+HP (AUC 0.91, 95\% CI 0.90-0.93).  Combining an image-only model result with other clinical data improves upon the clinical data-alone, but does not reach the performance of directly modeling all data (Table \ref{TableS12}).  

\begin{table}
\centering
\caption{Comparing the performance of models trained directly on different predictor sets, and models that ensemble image models with covariates.  Each primary model is a logistic regression model to predict fracture.  Naive Bayes ensembles were constructed to combine evidence from the image model and other predictor sets without knowing the interdependencies between them.}
\label{TableS12}
\resizebox{0.92\textwidth}{!}{\begin{tabular}{|l|l|l|l|l|l|l|l|l|l|l|l|l|}
\hline
Classifier & auc & auprc & threshold & specificity & sensitivity & accuracy & npv & ppv & tn & tp & fn & fp\\
\hline
pt & 0.79 (0.75-0.82) & 0.15 & 0.028 & 0.62 & 0.83 & 0.63 & 0.99 & 0.072 & 3,567 & 171 & 36 & 2,196\\
\hline
ptHp & 0.87 (0.85-0.89) & 0.14 & 0.033 & 0.75 & 0.89 & 0.76 & 0.99 & 0.115 & 4,339 & 185 & 22 & 1,424\\
\hline
imgPt & 0.86 (0.83-0.88) & 0.24 & 0.045 & 0.83 & 0.78 & 0.83 & 0.99 & 0.143 & 4,797 & 161 & 46 & 966\\
\hline
imgPtHp & 0.91 (0.90-0.93) & 0.40 & 0.048 & 0.85 & 0.83 & 0.85 & 0.99 & 0.168 & 4,910 & 172 & 35 & 853\\
\hline
nb\_imgPtHp & 0.90 (0.88-0.93) & 0.33 & 0.052 & 0.86 & 0.82 & 0.86 & 0.99 & 0.175 & 4,969 & 169 & 38 & 794\\
\hline
nb\_imgPt & 0.84 (0.81-0.87) & 0.22 & 0.034 & 0.83 & 0.77 & 0.82 & 0.99 & 0.138 & 4,764 & 160 & 47 & 999\\
\hline
\end{tabular}}
\end{table}

\section*{Discussion}
CNNs can use radiograph pixels to predict not only disease, but also numerous patient and hospital process variables.  Several prior studies demonstrated the ability of CNNs to learn patient traits\cite{Poplin2018-wz}, and image acquisition specifications\cite{Gale2017-ae,Madani2018-ok,Olczak2017-ob}, but to our knowledge this is the first comparative report on how deep learning can detect disease, demographics, and image acquisition targets, and use all of these factors to improve prediction performance.  We further show that CNNs learn to encode the statistical relationships between patient and hospital process covariates and hip fracture.  Despite CNNs directly encoding some of these variables, the direct addition of patient and hospital process variables in multimodal models further boosts predictive performance, while secondarily combining an image model with other variables is less beneficial.  This study expands on previous work that incorporates interdependencies between disease comorbidities to improve CNN predictions\cite{Yao2017-wt} by considering the compendium of patient and hospital process variables involved in routine clinical care.
\bigskip

\noindent The standard of care a patient receives is based on his or her differential diagnosis and the pre-test probability of disease, and these differences in diagnostic work-ups can induce structure into healthcare data that is learned by statistical learning algorithms.  We reproduced known associations between patient traits and hip fracture (e.g., older age, low body weight)\cite{Kanis2005-fb}.  Because of these known patient associations, middle-aged or elderly patients clinically suspected to have a hip fracture should have a follow-up MRI if the x-ray is negative or indeterminant.\cite{Ward2013-kr} The different standards of care based on patients’ global health was previously noted by Agniel et al. to be more predictive than the biological values that were being measured.\cite{Agniel2018-qu} They concluded that “Electronic Health Record (EHR) data are unsuitable for many research questions.”  
\bigskip

\noindent The ML model by Agniel et al. had discrete inputs that could be easily separated (e.g., the time a lab is ordered is a process and the protein measurement is biological).  In contrast, we have shown that DL can extract PT + HP variables from inseparable image pixels, so we used matched subsampling to alter the associations between fracture and PT + HP variables.  When deep learning rare conditions, it is common to perform class balancing by down-sampling normal examples, but this is generally done randomly without considering PT or HP variables.  Clinical trials with a case-control study design sometimes down-sample the normal patients with a matching routine to evenly distribute known confounding variables.\cite{Pearce2016-bs} After matching radiographs by patient and process factors, the associations between these factors and fracture were blotted out and the model was no longer able to detect hip fracture.  This loss of predictive performance indicates that the model was predicting fracture indirectly through these associated variables rather than directly measuring the image features of fracture.
\bigskip

\noindent While most DL models perform random class balancing, demographics are available in the largest publicly available medical image datasets and have been used for radiograph matching.\cite{Brestel2018-gp} However, we found that a richer set of matching variables was required to uncover the dependency on confounding variables.  Our model’s performance was consistent after matching demographics, and the dependence on other variables was only revealed when additionally controlling by patient symptoms and hospital process variables.  The addition of patient symptoms to demographics matching had the side effect of equating the odds of fracture on the top two devices, consistent with the hypothesis that patients are triaged differently which induces HP biases into the data.  Given that DL is able to detect so many variables from the radiographs, balancing just demographics does not suffice to reveal the classification mechanisms that were exposed after balancing with PT + HP variables.  
\bigskip
 
\noindent Although our model was dependent on covariates to predict fracture, the previously reported DL model for hip fracture detection by Gale et al. was not.  We attribute this to the fact that Gale et al. used different modeling strategies and disparate training data which had fewer covariates available for radiograph matching and weaker covariate-fracture associations.  The patient populations came from different clinical settings (Emergency Department [ED] patients in Australia versus a ED, inpatients, and outpatients at Mount Sinai Health System [MSHS] New York City).  There were stronger associations between fracture and HP covariates at MSHS, possibly because radiographs were collected from a wider range of clinical settings.  The metadata collected from MSHS was more extensive so patients were matched on additional symptoms and HP variables.  The labels for Gale et al. were semi-manually curated whereas MSHS was solely inferred from clinical notes (see limitations below).  Gale et al. used a customized DenseNet architecture that allowed them to maintain high resolution of the region of interest (1024 x 1024 pixels, in contrast to the 299 x 299 of the inception v3 model used for our primary experiments). In addition, Gale et al. developed a set of labels for the location of the hip joint and trained a preprocessing CNN to zoom in on this region of interest before applying their DenseNet classifier.  This sequential CNN localization and detection approach has been referred to as using cascaded CNNs. 
\bigskip

\noindent Since confounding variables have been shown to occur at the edge of whole radiographs (within the image, but outside of the patient), a pre-localization step may mitigate the impact of non-biological variables being considered by DL classification models.\cite{Zech2018-ok} Various localization strategies have been employed before applying DL to radiology data.  We and others perform whole image classification since labels can be extracted in a semi-automated fashion from clinical notes.\cite{Islam2017-ic, Kim2018-zf, Lakhani2017-ya, Rajpurkar2017-tx, Rajpurkar2017-po, Titano2018-kv, Yao2017-wt} Some investigators manually indicate the region of interest on every single image before “automated detection”\cite{Chung2018-uw} or use heuristics to crop scans\cite{Olczak2017-ob, Tomita2018-xx}.  The use of cascaded CNNs to initially segment images requires an additional training dataset, but it has been embraced by several investigators.\cite{Gale2017-ae, Roth2015-pc, Roth2016-gf, Shin2016-ul}  The description of the FDA-approved OsteoDetect suggests that it was not performing classification but trained with pixel-level data to perform pure segmentation.  The combination of zooming down to a region of interest and maintaining high resolution of the CNN’s receptive field helped ensure that valuable fracture-specific radiographic findings were not lost in the process of image downsampling. 
\bigskip

\noindent Deep learning is frequently criticized because the predictions lack clear attribution, and people are uneasy about blindly trusting a machine.  Our Naive Bayes experiment provides a more tangible reason we should seek to improve the interpretability of these models specific to the application of CAD.  If a DL model is acting fully autonomously, as has been proposed for retinopathy screening\cite{Abramoff2018-bi}, then it can benefit from incorporating patient and process variables, and it is inconsequential whether this behavior is explicitly known.  However, if the algorithm is intended to provide a radiologist with an image risk score so the radiologist can consider this in addition to the patient’s documented demographics and symptoms, a patient interview and physician exam, then it is undesirable if the CNN is unknowingly exploiting some portion of these data. This behavior creates uncertainty around how much of the CNN’s prediction is new evidence or redundant information.  If the clinician presumes an image-only interpretation model is not leveraging patient or process variables, they may consider the evidence as statistically independent, as the Naive Bayes model assumes, which produces a worse prediction than one based on full knowledge of interdependencies.  As clinicians are frequently considering other evidence sources not available to the model (e.g., clinical history, past imaging), it is important to enable them to interpret a model in the context of a patient’s larger clinical scenario.  
\bigskip

\noindent Previous image recognition CAD studies may not have considered PT and HP variables because they are inconsistently reported and/or unstructured in the EHR.  Better standards for documenting the risk factors a clinician considers during diagnosis could allow more image recognition studies to incorporate multimodal predictor sets.  
\bigskip

\noindent Several factors limit this study’s input data quality and model predictive performance.  First, we did not have gold standard images: plain film radiology is the first line of the diagnostic work-up, but not the gold standard imaging modality for hip fracture detection.\cite{Ward2013-kr} Second, our labels have limited accuracy: we used natural language processing to automatically infer the presence of fracture in a radiographic study from the clinical note.  Since the radiologist had multiple images and non-imaging data, a fracture may not be discernible on every image we labelled as fracture, which has been previously reported.\cite{Olczak2017-ob} Third, our covariate data have limited accuracy: we imputed missing patient BMIs and HP variables when they were not documented consistently.  Fourth, our preprocessing reduces image resolution: we used a pre-trained network which required us to downscale the images from full resolution to 299×299 pixels.  We further reduced the detail in the image representation by using a CNN feature dimensionality reduction to constrain the image feature space.  This simplification was done to expedite model training since our investigative questions involved training and comparing many models rather than training a best model.
\bigskip

\noindent Further research is needed to investigate sampling biases and generalization in DL observational medical datasets.  Pixel-level image annotation can enable pre-segmentation in cascaded networks.  Datasets with complete metadata can perform matched experimental designs, but the largest medical radiology datasets that are publicly available do not contain image acquisition specifications or hospital process variables, and we would need to develop more intricate methods to mitigate the influence of non-biological signal in these resources.  Genomic analyses have accounted for unmeasured confounding variables via surrogate variable analysis\cite{Leek2007-et}, factor analysis\cite{Stegle2012-wg} and mixed models\cite{Hoffman2016-pb}. DL approaches to mitigate undesired signal include adversarial networks\cite{Ganin2015-rl} and domain separation networks\cite{Bousmalis2016-kz}.  
\bigskip

\noindent DL algorithms can predict hip fracture from hip radiographs, as well as many patient and hospital process variables that are associated with fracture.  Observational medical data contain many biases, and radiographs contain non-biological signal that is predictive of disease but may not be ideal for computer-aided diagnosis applications.  Directly extending DL image models with known covariates can improve model performance, and performing localization steps before classification may mitigate dependence on these covariates.  Given that the largest public datasets lack covariate annotations, further research is needed to understand what specific findings are contributing to a model’s predictions and assess the impacts of DL’s incorporation of non-disease signal in CAD applications.

\section*{Methods}
\subsection*{Datasets}
We collected a new dataset from Mount Sinai Health System (MSHS), and then re-analyzed previously published results from the University of Adelaide.  The University of Adelaide data and deep learning model methods were previously described\cite{Gale2017-ae} and differ from those used on MSHS data.  Below we elaborate on the MSHS dataset and model development methods.  Separate models were developed from each site and used to predict fracture on internal test set images.  The patient matching and model evaluation described below were applied to both site’s test set predictions.  Train-test partition was stratified by patient so no patient had radiographs in both train and test sets.  

\subsubsection*{Imaging Studies}
23,602 hip radiographs were retrieved from the Picture Archiving Communication System (PACS) in DICOM file format, of which 23,557 radiographs from 9,024 patients were included in the study (see Figure \ref{fig:FigS1}).  The retrospective study protocol was reviewed by the IRB and designated exempt.  Hip radiographs were collected from 2008 to 2016.  Image were acquired for routine medical practice from several clinical sites (inpatient 4,183, outpatient 3,444, ED 7,929, NA 8,005) on 12 devices manufactured by Fujifilm, GE, Konica, and Philips.

\subsubsection*{Preprocessing of imaging data}
Radiographs were standardized to a common size and pixel intensity distribution.  Image were downsampled and padded to a final size of 299×299 pixels.  Pixel intensity mean and standard deviation were normalized per-image. 

\subsubsection*{Labels}
Image labels were parsed from two sources: the DICOM file header and clinical notes (see Figure \ref{fig:Fig1}B).  The DICOM file headers recorded the image acquisition specifications in a tabular format.  The clinical notes recorded patients’ demographics and radiologists’ interpretation times in tabular format, and the patients’ symptoms and radiologists’ image impression in free text.  These clinical notes were retrieved via Montage (Nuance Communications, Inc, New York, NY).  To abstract the presence of a symptom (i.e., pain, fall) we applied regular expressions to the noted indication.  To abstract fracture from the physicians’ image interpretation we used a word2vector based algorithm previously described by Zech et al.\cite{Zech2018-sd} Further label processing was performed to remove infeasible values and binarize values for binary classification models, as described in the supplementary methods.

\subsubsection*{Metadata Value Filtering}
We applied several feasibility filters to remove improbably data values.  For any values outside the expected range, we replaced the value with an `NA` indicator.  Entries with `NA` values were ignored during the primary regression and logistic modeling, and new values were imputed for secondary analyses.  In the clinical note database from Montage we retrieved the latency time between image order, acquisition, and interpretation.  We removed any value that was less than one minute, or greater than one day.  Patient BMI values documented greater than 60 were removed.

\subsubsection*{Target Binarization}
For logistic regression models, we coerced continuous or categorical variables into a binary representation.  For continuous variables, we simply took the median value and indicated whether values were greater than the median or not.  For categorical variables, we used a natural abstraction where possible or took the two most common levels and filling in the remaining entries with `NA` values.  For day of week, we abstracted to weekday and weekend.  For scan projection, we abstracted to lateral versus bilateral views.  For all other categorical variables we kept the two most common values and removed the others.  

\subsubsection*{Metadata Value Imputation}
To train multimodal models and perform radiograph case-control matching, we needed to handle missing data in numerous PT and HP variable fields.  For categorical variables, missing entries were replaced by an explicit “(Missing)” value.  The only PT variable with missing data was BMI.  To impute BMI we we trained linear regression models on the subset of data with available BMIs using each combination of predictor variables, possibly with imputed HP variables.  We used the model with all predictor sets and imputed HP variables to impute all missing BMI entries.  For other continuous variables, we simply performed median imputation.

\subsection*{Model Architecture and Training}
\subsubsection*{Image Model Architecture}
We used deep learning models called Convolutional Neural Networks (CNNs) to compute abstract image features from input image pixel arrays.  Deep learning models require an abundance of training images to learn meaningful image features during an initial training phase to select parameters that improve a model’s performance on a particular task.  We used the inception-v3 CNN architecture\cite{Szegedy2015-uz} with parameters that have been optimized for natural object recognition in the ImageNet challenge\cite{Russakovsky2014-vi}.  We use the pre-trained model to encode radiograph image features and then re-train the final layer, which is a practice called transfer learning and has previously been performed for image recognition tasks in medical radiology\cite{Chung2018-uw, Kim2018-zf, Olczak2017-ob}.  The final classification layer is removed, and we compute the penultimate layer of 2048 image feature scores for each radiograph.  We use these abstract feature vectors in subsequent unsupervised models.  Deep learning processing was performed with the python packages keras and tensorflow.

\subsubsection*{Unsupervised Analysis t-Distributed Stochastic Neighbor Embedding (t-SNE)}
After computing the image features for each image, we use several dimensionality reduction techniques to visualize the distribution of image variation.  For comparison, we featurized images with one inception-v3 model with randomly initialized parameters and a second that was pre-trained on ImageNet.  We performed t-Distributed Stochastic Neighbor Embedding (t-SNE) to project the image feature vector into a 2d plane with the R package Rtsne (initial PCA to 50 dimensions, perplexity 30, theta 0.5, initial momentum 0.5, final momentum 0.8, learning rate 200).  

\subsubsection*{Supervised Analysis}
We fine-tune models using image and/or covariate explanatory variable sets and predict binary and continuous outcome variable.  For image-only models, we use the 10 principal components of the 2048-dimensional image feature vectors as model input (which describes 68\% of image variation in the 2048-d space), and for combined image-metadata models we concatenate the 10-principal component image vector with the scalar metadata values.  To predict binary variables, we use logistic regression fit to maximize AUC and for continuous variables linear regression to minimize RMSE.  This fine-tuning was done in R using the caret package.\cite{Kuhn2008-uo}

\subsubsection*{Naive Bayes}
Naive Bayes was used to ensemble the predictions of a model based on patient demographics and symptoms and a model that only uses the image.  The prior probability was estimated using a kernel based on 10-fold cross validation with the training partition.  The R package klaR was used to compute the posterior probability of fracture assuming independence between predictors.

\subsection*{Statistical Methods}

\subsubsection*{Case-Control Matching}
After computing the image features for each image, we use several dimensionality reduction techniques to visualize the distribution of image variation.  For comparison, we featurized images with one inception-v3 model with randomly initialized parameters and a second that was pre-trained on ImageNet.  We performed t-Distributed Stochastic Neighbor Embedding (t-SNE) to project the image feature vector into a 2d plane with the R package Rtsne (initial PCA to 50 dimensions, perplexity 30, theta 0.5, initial momentum 0.5, final momentum 0.8, learning rate 200).  

\subsubsection*{Classifier Receiver Operating Curves}
All Receiver Operator Curves (ROC) analyses and comparisons were done with the R pROC package.\cite{Robin2011-we} We compute Area Under the Curve (AUC) with 95\% confidence intervals determined by 2000 bootstrapped replicates stratified to control case-control frequency.  Comparisons between ROC curves were done with unmatched test cohorts when modeling different outcomes and cohorts and were therefore performed by bootstrapping 2000 stratified replicates with a two-sided test.  The exceptions are the fracture detection models which used different predictor sets for the same cohort (e.g., image-only, hospital process predictors, multimodal, and Naive Bayes analysis).  In these cases the test group was correlated for all models and we used Delong’s asymptotically exact method to compare AUCs in a two-sided test.  We select the best operating point with Youden’s method to further compute sensitivity, specificity, and other threshold-dependent statistics.  

\subsubsection*{Odds Ratios}
Associations between each variable and fracture are performed with the binarized covariates and a two-sided Fisher’s exact test for count data.  

\subsection*{Acknowledgements}
This work was supported by Verily Life Sciences, LLC as part of the Verily Academic
Partnership with Icahn School of Medicine at Mount Sinai, and by the National Institutes of
Health, National Center for Advancing Translational Sciences (NCATS), Clinical and
Translational Science Award.
\bigskip

\noindent We thank Drs. Eric Oermann and Joseph Titano for contributing the dataset from Mount Sinai
Hospital, and Lyle J. Palmer, Gustavo Carneiro, and Andrew Bradley for contributing the test
dataset from The University of Adelaide.

% Use the Nature BiBTeX style
\bibliographystyle{naturemag}

\end{document}